\documentclass{article}

\usepackage{arxiv}

\usepackage[utf8]{inputenc} 
\usepackage[T1]{fontenc}    
\usepackage{hyperref}       
\usepackage{url}            
\usepackage{booktabs}       
\usepackage{amsfonts}       
\usepackage{nicefrac}       
\usepackage{microtype}      
\usepackage{lipsum}
\usepackage{graphicx}
\usepackage{tabularx}
\usepackage{multirow}
\usepackage{enumitem}
\usepackage{amsmath,amsfonts,amssymb,amsthm,bm}
\usepackage{color}
\usepackage{algorithm}
\usepackage{algpseudocode}
\usepackage{subcaption}

\graphicspath{ {./ } }

\title{DuDGAN: Improving Class-Conditional GANs \\ via Dual-Diffusion}

\author{
 Taesun Yeom \\
  School of Mechanical Engineering\\
  Chung-Ang University\\
  \texttt{taesun0415@cau.ac.kr} \\
   \And 
 Minhyeok Lee \\
  School of Electrical and Electronics Engineering\\
  Chung-Ang University\\
  \texttt{mlee@cau.ac.kr} \\
}

\begin{document}
\maketitle
\begin{abstract}
Class-conditional image generation using generative adversarial networks (GANs) has been investigated through various techniques; however, it continues to face challenges such as mode collapse, training instability, and low-quality output in cases of datasets with high intra-class variation. Furthermore, most GANs often converge in larger iterations, resulting in poor iteration efficacy in training procedures. While Diffusion-GAN has shown potential in generating realistic samples, it has a critical limitation in generating class-conditional samples. To overcome these limitations, we propose a novel approach for class-conditional image generation using GANs called DuDGAN, which incorporates a dual diffusion-based noise injection process. Our method consists of three unique networks: a discriminator, a generator, and a classifier. During the training process, Gaussian-mixture noises are injected into the two noise-aware networks, the discriminator and the classifier, in distinct ways. This noisy data helps to prevent overfitting by gradually introducing more challenging tasks, leading to improved model performance. As a result, our method outperforms state-of-the-art conditional GAN models for image generation in terms of performance. We evaluated our method using the AFHQ, Food-101, and CIFAR-10 datasets and observed superior results across metrics such as FID, KID, Precision, and Recall score compared with comparison models, highlighting the effectiveness of our approach.
\end{abstract}

\begin{figure}[t]
    \centerline{\includegraphics[width=\columnwidth]{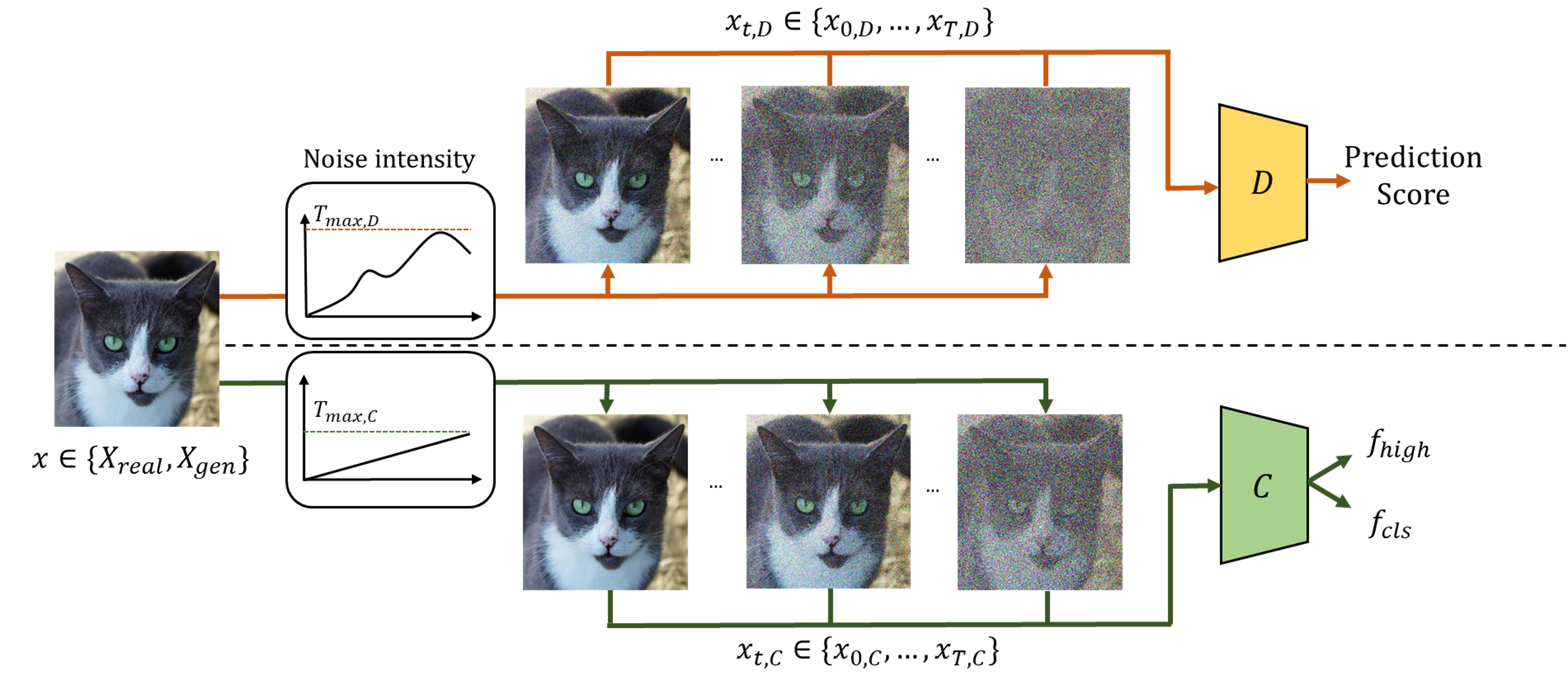}}
    \caption{{\bf Structural Overview of DuDGAN.} An arbitrary input image is first sampled from the training dataset, which is then diffused by an independent noise schedule for both the discriminator and the classifier. Subsequently, the generator produces high-quality and class-specific images with the aid of the supervision of the discriminator and classifier.}
    \label{figure:figure_1}
\end{figure}
\section{Introduction}
Generative adversarial networks (GANs) and their numerous variations have demonstrated significant success within the realm of computer vision. These networks have shown impressive performance in a wide array of tasks, such as image generation \cite{karras2017progressive,brock2018large,lee2019controllable,sauer2021projected,choi2020stargan}, image-to-image translation \cite{zhu2017unpaired,choi2018stargan,emami2020spa,xie2021unaligned,ko2023superstargan}, video generation \cite{tulyakov2018mocogan,wang2020imaginator,chu2020learning}, 3D reconstruction \cite{schwarz2020graf,meng2021gnerf,niemeyer2021giraffe,chan2022efficient}, and GAN inversion \cite{richardson2021encoding,tov2021designing,wang2022high}. The field of image generation, in particular, has experienced considerable advancements in both quality and diversity, largely attributed to the development of style-based architectures \cite{karras2019style,karras2020analyzing,karras2020training,karras2021alias}.

Typically, image generation using GANs can be classified into two categories: unconditional and conditional image generation. While unconditional image generation does not require any additional information, the conditional approach necessitates supplementary input, such as a specific image or class label. A majority of conditional GAN models aim to control the output image through auxiliary supervision during the training process. Consequently, numerous studies, including \cite{mirza2014conditional,odena2017conditional, lee2019controllable,kang2020contragan}, have been conducted to enhance the quality of generated images. Despite the notable results achieved thus far, conditional image generation remains more challenging due to the need for learning over smaller intra-class data distributions. Additionally, these methods are hindered by the necessity of a vast, labeled dataset and plenty of iterations to ensure stable training.

Nevertheless, the process of collecting and curating a large, class-specific dataset is both labor-intensive and time-consuming. Moreover, in this case, conditional GANs often encounter several issues during the training phase, such as mode collapse and gradient explosion problems \cite{shahbazi2022collapse, kang2021rebooting,tseng2021regularizing}. Consequently, it is crucial to explore suitable techniques for training conditional GANs with limited data.

While some data-efficient GAN training methods have been proposed \cite{Zhao2020Differentiable, karras2020training,wang2022diffusion}, these approaches predominantly focus on training within an unconditional data regime rather than using class-labeled images. Indeed, some recent research has aimed to enhance conditional image generation with small datasets. For instance, Transitional-CGAN \cite{shahbazi2022collapse} introduced a novel training strategy that combines unconditional and conditional training to address condition-induced mode collapse. However, this method primarily concentrates on reducing supervision for conditions during the early training stage and thus may not be an effective solution for preventing collapse in later stages of training. Moreover, this approach is not efficient in terms of iteration efficacy due to the extensive scales involved in the transitional process.

In response to these challenges, we present DuDGAN (Fig.~\ref{figure:figure_1}), a robust method for class-conditional image generation that excels in iteration-efficient training. Our approach comprises three distinct networks: a generator, a discriminator, and a classifier. Drawing inspiration from previous work \cite{wang2022diffusion} that trains a discriminator using noise injection, our objective is for a timestep-dependent classifier to learn and output class-conditional information during training while incorporating a diffusion-based noise injection process. Concurrently, the timestep-dependent discriminator acquires prior knowledge from the classifier to discern whether images are real or fake.

The classifier's output consists of two types: high-dimensional class information for calculating contrastive loss \cite{khosla2020supervised} and class-dimensioned logits for classification loss. Throughout the training process, we employ an appropriate diffusion intensity for both the discriminator and the classifier, determined by each network's degree of overfitting. Consequently, our method generates images with high intra-class variation and effectively prevents mode collapse. 

Our key contributions are as follows:
\begin{itemize}
    \item We investigate the impact of using an additional classifier trained with a diffusion-based noise injection process for class-conditional image generation.
    \item We propose a novel approach termed \textit{dual-diffusion}, which signifies the collaboration between the discriminator and the classifier, both of which are trained using diffusion-based noise injection.
    \item DuDGAN achieves fast convergence within a limited number of iterations, thereby accomplishing both high-quality generation and iteration-efficient training.
    \item As a result, DuDGAN achieves superior performance in compared to state-of-the-art GAN models on the AFHQ \cite{choi2020stargan}, Food-101 \cite{bossard2014food}, and CIFAR-10 \cite{krizhevsky2009learning}.
\end{itemize}

\section{Related Work}
\label{sec2}
{\bf Training generative adversarial networks with class-conditional images.}\; GANs \cite{goodfellow2020generative} are generative models designed to produce realistic data by approximating a real data distribution $p(x)$. Two primary neural networks, the discriminator and the generator, undergo simultaneous training to achieve their objectives. The discriminator learns to differentiate between real and fake data, while the generator strives to generate data that can deceive the discriminator. In this context, the objective function for Vanilla GAN \cite{goodfellow2020generative} can be expressed as:
\begin{equation}
    \min _G \max _D V(G, D)=E_{x \sim p(x)}[\log (D(x))]+E_{z \sim p(z)}[\log (1-D(G(z)))],
\end{equation}
where $z\sim p(z)$ is a noise vector from a particular distribution (e.g., Gaussian distribution) and $x\sim p(x)$ is sampled from the real data distribution. Under ideal conditions, the discriminator outputs a probability of one-half for any given input. 

However, unconditional GAN models are unable to generate the desired images as they train over the entire data distribution, regardless of class-wise information. To address this issue, Mirza and Osindero introduced CGAN \cite{mirza2014conditional}, which generates conditional images by incorporating a class label into both the generator and discriminator. The basic form of the objective function for conditional GANs with discrete conditional information c is as follows:
\begin{equation}
    \min _G \max _D V(G, D)=E_{x \sim p(x)}[\log (D(x,c))]+E_{z \sim p(z)}[\log (1-D(G(z,c)))].
\end{equation}

Several studies have been conducted in this area. ACGAN \cite{odena2017conditional} enhances the performance of conditional image generation by employing an auxiliary classifier to output class information for backpropagation. Transitional-CGAN \cite{shahbazi2022collapse} uses a linear transition function during the training phase to prevent mode-collapse in conditioning. Rebooting-ACGAN \cite{kang2021rebooting} projects a vector onto a hypersphere to mitigate mode collapse caused by gradient explosion.  

\noindent{\bf Diffusion-based generative models.}\; Particularly within the domain of computer vision, the diffusion model is considered the general form of denoising diffusion probabilistic models (DDPM) \cite{ho2020denoising}. It consists of a two-way Markov chain, known as the forward and reverse processes. In the forward process, Gaussian noise is gradually injected into the data at discrete timesteps $t \in\{0,1, \ldots, T-1, T\}$. As a result, the data becomes random noise $\mathcal{N}$(0,1) after the final $T$ steps. Note that the predefined variance schedule $\beta_t$ and variance $I$ in the equations below do not have any learnable parameters. The equation for the forward noising process is as follows: 
\begin{equation}
F\left(x_t \mid x_{t-1}\right):=\mathcal{N}\left(x_t ; \sqrt{1-\beta_t} x_{t-1}, \beta_t I\right), \hspace{0.5cm} \beta_t\in(0,1).
\end{equation}

In contrast, the reverse process entails denoising the data from noise to target data. This process is governed by a set of parameters $\theta$. The reverse process can be represented as the model of latent variables:
\begin{equation}
R_\theta\left(x_{t-1} \mid x_t\right):=\mathcal{N}\left(x_{t-1} ; {\mu}_\theta\left(x_t, t\right), {\Sigma}_\theta\left(x_t, t\right)\right).
\end{equation}

The set of parameters $\theta$ at each denoising step can be calculated by parameterizing a specific neural network within the model. Diffusion-GAN \cite{wang2022diffusion} demonstrates that diffusion-based data augmentation is effective for mode-catching and provides non-leaking augmentation for the discriminator. In this paper, we explore the efficacy of the forward noise injection process in improving the quality of image generation in a class-conditional setting.

\section{Method}
\subsection{Noise injection through forward diffusion process}
The primary training objective of a class-conditional GAN is to generate high-quality and photorealistic conditional samples by training through the real data distribution $p_x$, while predicting modes over a limited class-wise distribution $p_{x|c}$, which is a subset of $p_x$. In this process, Gaussian noise is injected into both the discriminator's and classifier's inputs using a forward diffusion chain. As mentioned in Sec.~\ref{sec2}, a distribution component derived from the noise injection process of an arbitrary noisy sample $x_j\in\{x_0, \ldots, x_T\}$ through the forward Markov chain can be expressed as \cite{ho2020denoising,sohl2015deep}:
\begin{equation}
F_t(x_j|x_0)=\mathcal{N}(x_j;\sqrt{\bar{\alpha}_t}x_0,(1-\bar{\alpha}_t)\sigma^2I),
\end{equation}
where the distribution component depends on the timestep variable $t$. In the equation, ${\bar{\alpha}}_t:=\prod_{k=1}^{t}{1-\beta_k}$, and $x_0$ is the real or generated image that is not perturbed by the Gaussian noise. Furthermore, by applying the reparameterization trick \cite{ho2020denoising}, the noisy sample $x_j$ can be summarized as the linear combination of original data and noise:
\begin{equation}
    x_j(x_0, j)=\sqrt{{{\bar{\alpha}}_j}} x_0+\sqrt{1-\bar{\alpha}_j} \sigma \epsilon, \hspace{0.5cm} j \in \{0,1, \ldots, T-1, T\}.
\end{equation}

In discrete timestep $t$, Gaussian noises are injected into real or generated images by the equation. As the timestep increases, more information loss occurs in the sample. Each timestep-dependent distribution forms a Gaussian-mixture distribution for an arbitrary timestep.

However, since our method focuses on training with class-conditional images, each image is affected by a conditional mixture distribution defined as the summation of mixture weight $w_t$ under maximum timestep $T$ and class $c$ \cite{wang2022diffusion}:
\begin{equation}
    F_c\left(x_j \mid x_0\right)_{x_0 \sim p_{x \mid c}}:=\sum_{t=1}^T\left\{w_t \cdot F_t\left(x_j \mid x_0\right)\right\}.
\end{equation}

Thus, we can sample a noisy image $x_j$ by sampling the timestep $t$ from the mixture component.

\begin{figure}[t]
\centerline{\includegraphics[width=\columnwidth]{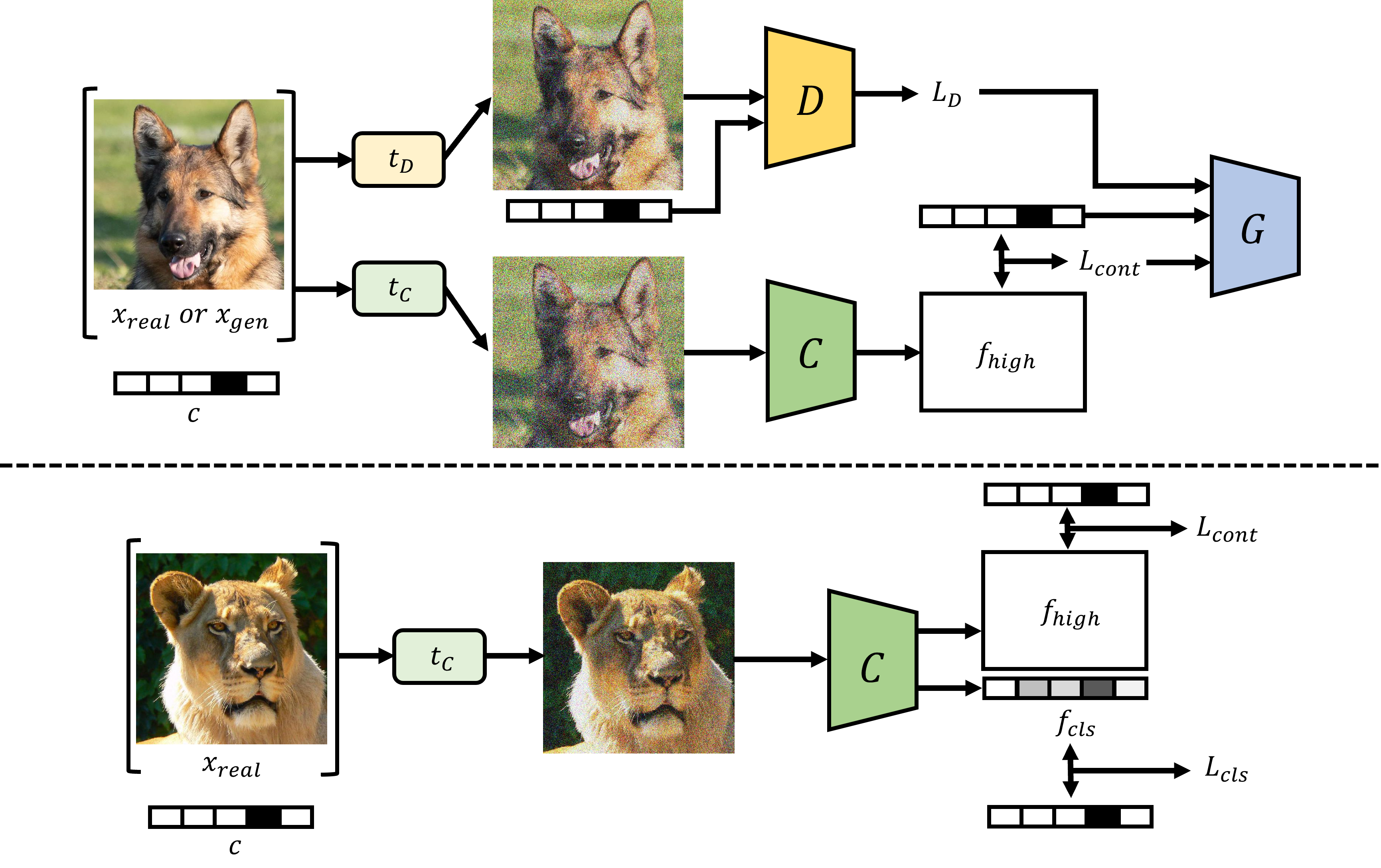}}
    \caption{{\bf DuDGAN Training Procedure.} The procedure has two components: (1) losses used during generator training, which include the discriminator's prediction score and a contrastive loss in a high-dimensional feature space with original class labels, and (2) independent training of the classifier, including both a contrastive loss and cross-entropy for classification loss. Input images undergo a separate Gaussian-mixture noise injection process, causing information loss, which is accounted for by a predefined parameter for high-quality generative results.}
    \label{Figure 2.}
\end{figure}

\subsection{Additional Classifier for Conditional Image Generation}
To achieve high-fidelity and diversity of generated images, models for class-conditional image generation must have the ability to handle extensive class-wise distribution as much as whole distribution. This necessitates an additional network that deals with class information. Inspired by previous work \cite{ko2023superstargan}, which demonstrates the effectiveness of an independent classifier network intending to increase class-wise and class-aware capacity for GAN training, our method includes an independent classifier that receives real or generated images with Gaussian-mixture noise and outputs class information. Note that the classifier input consists only of the real or generated image and does not contain class labels. Consequently, the classifier can predict distribution beyond the bounded information configured by class-wise images for training. 

This procedure prevents overfitting on the training set and enables learning broadly of the class information. Furthermore, classifier outputs comprise two-level conditional information represented as $f_{high}$ and $f_{cls}$. $f_{high}$ consists of a high-dimensional latent code that contains high-frequency class-conditional features, while $f_{cls}$, class logits for domain classification on the class of the input image and the class predicted by the network, are formed of a vector whose dimension is the same as class labels. For accurate training on the classifier, the classifier is trained only with real images, not with images generated by the generator. With an arbitrary noisy image $x_j$, classifier outputs can be written as:
\begin{equation}
    (f_{\text {high }}, f_{c l s})=C\left(x_j\right).
\end{equation}

\subsection{Dual-Diffusion Process}
Our method targets training two neural networks simultaneously, the discriminator and the classifier, through an independent diffusion-based noise injection process. For both networks, our approach is based on \cite{wang2022diffusion}, which employs the procedure of gradually presenting the discriminator with a more challenging task by first showing clear samples and then introducing noisy samples.

The discriminator, which undertakes the bi-classification task of predicting the realness score by taking real and generated images as input during the training process, aims to self-supervise the noise intensity by leveraging a predefined hyperparameter $r_d$, which indicates the extent to which the discriminator is overfitted to the training set \cite{karras2020training,wang2022diffusion}. Considering that the intensity of noise is determined by the discrete timestep $t$, the process of iteration $k$, which is a multiple of 4, is summarized as follows: 
\begin{equation}
    T_{k, D}=T_{k-4, D}+\operatorname{sign}\left(r_d-D_{\text {target }}\right) * \operatorname{const}.  \label{eq9}
\end{equation}
Note that $T_{k,D}\in(0,1)$ represents the maximum intensity of the noise injection process in iteration $k$, and $r_d$ is determined to be 0.6 by the experiment in \cite{karras2020training}.

The classifier aims to perform classification by labels according to the input image. Similarly to Eq. \ref{eq9}, the classifier receives a noisy sample with an independent noise schedule. To improve classification, we predefined the noise level for each iteration by dividing the total number of predefined training iterations $k_{\max}$. This can be interpreted as a linear increase in noise intensity from the original image to the fully noised image in proportion to the number of iterations. Furthermore, we bound the maximum diffusion intensity in training the classifier for better classification. The noise intensity for the independent classifier is written as follows:
\begin{equation}
    T_{k, C}=T_{k-4, C}+\frac{4}{k_{\max }},\hspace{0.5cm} T_{k,C}\in(0,0.3). \label{eq10}
\end{equation}
As in Eq. \ref{eq9} and Eq. \ref{eq10}, the diffusion intensities of the discriminator and the classifier are updated every 4 iterations. More details in this section are in the \textit{Supplementary Materials}.

\subsection{Overall Training with Diffusion}
The outline of the training procedure of our model is displayed in Fig.~\ref{Figure 2.}. To enhance quality and prevent collapse in class-conditional image generation, we propose a new form of overall loss functions. Three different networks, the generator, the discriminator, and the classifier, are jointly trained with loss functions to achieve their objectives. First, during the classifier's training with real images, $L_{cont}^{real}$ and $L_{cls}^{real}$ are calculated from the two-level outputs, $f_{high}$ and $f_{cls}$, respectively. $L_{cont}^{real}$ represents the supervised contrastive loss \cite{khosla2020supervised} derived from the high-dimensional latent space, while $L_{cls}^{real}$ denotes the simple classification error between the predicted and given labels.

Aiming to produce photorealistic and diverse images within class-wise distribution, the generator receives additional information from the classifier for the generated images. Thus, similar to the classifier, the generator's loss function consists of the contrastive loss of generated images, which guides the generator to produce high-fidelity images, while the original loss $L_G^{gen}$ remains.

Finally, the discriminator does not receive any informative gradient from the classifier, so the loss function remains the same as in the baseline model \cite{wang2022diffusion}, which is the non-saturating GAN loss. Summarizing this section, the following loss functions constitute our full training objective:
\begin{align}
    L_C & =\lambda_C \cdot L_{\text {cont }}^{\text {real }}\left(f_{\text {high }}, c_r\right)+\left(1-\lambda_C\right) \cdot L_{\text {cls }}^{\text {real }}, \label{eq.11}\\
    L_G & =\lambda_G \cdot L_G^{\text {gen }}+\left(1-\lambda_G\right) \cdot L_{\text {cont }}^{\text {gen }}\left(f_{\text {high }}, c_f\right), \label{eq.12}\\
    L_D & =L_{D}^{NS}, \label{eq.13}
\end{align}
where $\lambda_C$ and $\lambda_G$ are hyperparameters to modulate the training of the classifier and generator, respectively.

\section{Experiments}
\subsection{Experiment setup}
{\bf Datasets.}\; For class-conditional image generation with GANs, the dataset for training must contain label information. In this regard, we train and evaluate our method using three different datasets, each with a different resolution. The preprocessing steps and specifics for each dataset are described below:
\begin{itemize}
    \item AFHQ (512$\times$512) \cite{choi2020stargan}: AFHQ is a dataset originally consisting of three different categories: dogs, cats, and wild animals. To demonstrate the effectiveness of our method across various domains, we use a recreated version of the dataset \cite{ko2023superstargan}, which increases the number of classes from 3 to 7.
    \item Food-101 (128$\times$128) \cite{bossard2014food}: Food101 contains 101 different categories of food, where each class consists of 1k different images. We use a portion of the dataset consisting of 20 labels, without reducing class-wise data size. Additionally, due to the variability of the size of images, we preprocess each training image to 128$\times$128 size.
    \item CIFAR-10 (32$\times$32) \cite{krizhevsky2009learning}: CIFAR-10 is divided into 10 classes, each containing 50k training images and 10k test images.
\end{itemize}
{\bf Evaluation metrics.}\; To evaluate our method and compare it with other models, we employ Fr\'{e}chet inception distance (FID) \cite{heusel2017gans} and kernel inception distance (KID) \cite{binkowski2018demystifying} to measure the generation quality and assess whether the generation adheres to the distribution of the training data. Additionally, we utilize the Precision and Recall score \cite{kynkaanniemi2019improved} to gauge the fidelity and diversity of the generated samples. 

{\bf Implementation details.}\; To demonstrate that our method exhibits strength in fast convergence, all models are trained until the discriminator processes 10,000k images, a 60$\%$ smaller than those used in experiments with comparative models. Furthermore, especially for the classifier, we adopt AdamW optimizer \cite{loshchilov2017decoupled} instead of Adam optimizer \cite{kingma2014adam} to prevent class-induced overfitting. More implementation and training details can be found in the \textit{Supplementary Materials}. 

{\bf Comparison Models.}\; For fair comparison, our main experiment is built upon comparison with three different baseline methods, which are based on StyleGAN2-ADA \cite{karras2020training}: 1) class-conditional training of StyleGAN2-ADA (CStyleGAN2-ADA), 2) class-conditional training of Diffusion-StyleGAN2 (CDiffusion-GAN) \cite{wang2022diffusion}, and 3) default setting of Transitional-CGAN \cite{shahbazi2022collapse}. Note that, while CDiffusion-GAN is one of the baselines, the model is first introduced in this study, as a conditional version of Diffusion-GAN.

\subsection{Experimental result}
 {\bf Quantitative result.}\; As demonstrated in Table \ref{Table 1.}, our method surpasses the comparison models with respect to FID on the AFHQ and CIFAR-10 datasets, indicating superior generation quality. In particular, on the CIFAR-10 dataset, our model outperforms all other models, including the main baseline, CDiffusion-GAN, across all datasets. Notably, in the case of the AFHQ dataset, FID is reduced by 4.0\%. Furthermore, our method attains the highest Recall score on the CIFAR-10 dataset, signifying enhanced diversity in the generated samples. Although our method exhibits a marginally lower Precision and Recall score on the AFHQ dataset, it remains competitive with the top-performing model. 
 
\begin{figure}
    \centerline{\includegraphics[width=\columnwidth]{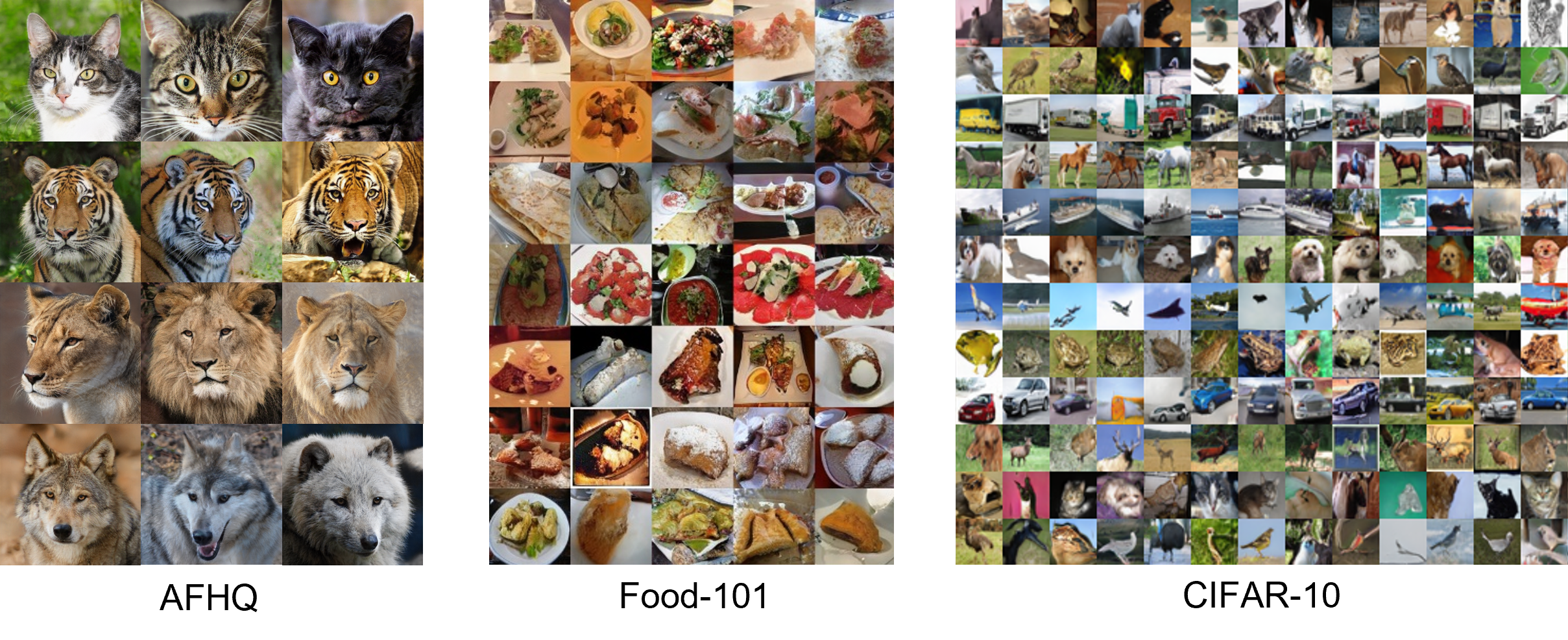}}
    \caption{{\bf Qualitative Result on AFHQ, Food-101, and CIFAR-10 Datasets.}}
    \label{Figure 3.}
\end{figure}

{\bf Qualitative result.}\; In addition to the quantitative results, we examine the visual quality of the generated images by our method. As shown in Fig.~\ref{Figure 3.}, the generated samples exhibit photo-realistic characteristics, demonstrating the efficacy of our approach. The images possess fine details, accurate colors, and clear textures, which contribute to their overall photo-realistic appearance. These results further validate the superiority of our method in generating high-quality, diverse, and visually appealing images within class-wise distributions. \\

\begin{table}[t]
\vspace{-0.1cm}
\centering
\resizebox{\columnwidth}{!}{%
\begin{tabular}{lcccccccccccc}
\hline
\multirow{3}{*}{\textbf{Method}} &
  \multicolumn{4}{c}{\textbf{AFHQ}} &
  \multicolumn{4}{c}{\textbf{Food-101}} &
  \multicolumn{4}{c}{\textbf{CIFAR-10}} \\
 &
  \multicolumn{4}{c}{(512 $\times$ 512)} &
  \multicolumn{4}{c}{(128 $\times$ 128)} &
  \multicolumn{4}{c}{(32 $\times$ 32)} \\
 &
  FID$\downarrow$ &
  KID$\downarrow$ &
  Pr$\uparrow$ &
  Re$\uparrow$ &
  FID$\downarrow$ &
  KID$\downarrow$ &
  Pr$\uparrow$ &
  Re$\uparrow$ &
  FID$\downarrow$ &
  KID$\downarrow$ &
  Pr$\uparrow$ &
  Re$\uparrow$ \\ \hline
CStyleGAN2-ADA &
  5.11 &
  \textbf{0.0010} &
  \textbf{0.75} &
  \textbf{0.31} &
  13.29 &
  0.0067 &
  0.57 &
  \textbf{0.18} &
  3.81 &
  0.0011 &
  \textbf{0.64} &
  0.56 \\
CDiffusion-GAN &
  5.31 &
  \textbf{0.0010} &
  0.64 &
  0.28 &
  21.07 &
  0.0130 &
  0.64 &
  0.09 &
  3.77 &
  0.0011 &
  0.63 &
  0.57 \\
Transitional-CGAN &
  7.80 &
  0.0011 &
  0.63 &
  0.20 &
  \textbf{10.37} &
  \textbf{0.0034} &
  0.63 &
  0.17 &
  4.25 &
  0.0013 &
  \textbf{0.64} &
  0.52 \\
Ours &
  \textbf{5.10} &
  \textbf{0.0010} &
  0.68 &
  0.29 &
  10.71 &
  0.0051 &
  \textbf{0.73} &
  \textbf{0.18} &
  \textbf{3.73} &
  \textbf{0.0009} &
  \textbf{0.64} &
  \textbf{0.58} \\ \hline
\end{tabular}%
}
\vspace{0.2cm}
\caption{{\bf Quantitative Result on AFHQ, Food-101, and CIFAR-10 Datasets.}}
\label{Table 1.}
\end{table}

\begin{table}[t]
\centering
\begin{tabular}{lcccc}
\hline
\multirow{3}{*}{\textbf{Experiment of $L_C$}}                              & \multicolumn{4}{c}{\textbf{CIFAR-10}}                               \\
                                                                           & \multicolumn{4}{c}{(32 $\times$ 32)}     \\
                                                                           & FID$\downarrow $ & KID$\downarrow $ & Pr$\uparrow $ & Re$\uparrow $ \\ \hline
$L_C=L_{\text {cont }}^{\text {real }}\left(f_{\text {high }}, c_r\right)$ & 68.14            & 0.0315           & 0.63              & 0.01          \\
$L_C=L_{\text {cls }}^{\text {real }}$                                     & 3.96             & 0.0011           & 0.63              & 0.57          \\
$\bm{L_C=\lambda_c \cdot L_{\text {cont }}^{\text {real }}\left(f_{\text {high }}, c_r\right)+\left(1-\lambda_c\right) \cdot L_{\text {cls }}^{\text {real }}}$ &
  \textbf{3.73} &
  \textbf{0.0009} &
  \textbf{0.64} &
  \textbf{0.58} \\ \hline
\end{tabular}%
\vspace{0.2cm}
\caption{\bf{Ablation Study: Classifier Loss Formulations.}}
\label{Table 2.}
\end{table}

\begin{table}[ht]
\vspace{-0.2cm}

\centering
\begin{minipage}[b]{0.47\textwidth}
\centering

\begin{tabular}{lcccc}
\hline
\multirow{3}{*}{\textbf{\begin{tabular}[c]{@{}l@{}}Experiment of $\lambda_G$\ \\ ($\lambda_C=0.95$)\end{tabular}}} \hspace{-1.5cm} & \multicolumn{4}{c}{\textbf{CIFAR-10}} \tabularnewline
& \multicolumn{4}{c}{(32$\times$32)} \tabularnewline
& FID$\downarrow$ & KID$\downarrow$ &Pr$\uparrow$ & Re$\uparrow$ \tabularnewline \hline
$\lambda_G=0.5$ & 4.55 & 0.0015 & 0.63 & 0.53 \tabularnewline
$\lambda_G=0.8$ & 54.88 & 0.0217 & \bf0.67 & 0.02 \tabularnewline
\bm{$\lambda_G=0.95$} & \bf3.73 & \bf0.0009 & 0.64 & \bf0.58 \tabularnewline \hline

\end{tabular}
\vspace{0.2cm}
\caption{\bf{\small Ablation Study: Variations in $\lambda_G$.}}
\label{Table 3.}
\end{minipage}
\hfill
\begin{minipage}[b]{0.47\textwidth}
\centering

\begin{tabular}{lcccc}
\hline
\multirow{3}{*}{\textbf{\begin{tabular}[c]{@{}l@{}}Experiment of $\lambda_C$\ \\ ($\lambda_G=0.95$)\end{tabular}}} \hspace{-1.5cm} & \multicolumn{4}{c}{\textbf{CIFAR-10}} \tabularnewline
& \multicolumn{4}{c}{(32$\times$32)} \tabularnewline
& FID$\downarrow$ & KID$\downarrow$ &Pr$\uparrow$ & Re$\uparrow$ \tabularnewline \hline
$\lambda_C=0.5$ & 4.02 & 0.0012 & 0.63 & 0.56 \tabularnewline
$\lambda_C=0.8$ & 4.24 & 0.0013 & \bf0.64 & 0.54 \tabularnewline
\bm{$\lambda_C=0.95$} & \bf3.73 & \bf0.0009 & \bf0.64 & \bf0.58 \tabularnewline \hline

\end{tabular}
\vspace{0.2cm}
\caption{\bf{\small Ablation Study: Variations in $\lambda_C$.}}
\label{Table 4.}
\end{minipage}

\end{table}

\subsection{Ablation Study}
In the ablation study, we conduct two distinct experiments by modifying the formulation of the classifier's loss function and adjusting the hyperparameter $\lambda$ in both the generator and the classifier. The experiments in this section are based on the class-conditional training of the CIFAR-10 dataset.

{\bf Two-Level Output of the Classifier.}\; Our method's classifier generates a two-level loss, computed as the label-dimensioned logits and contrastive loss, with the aim of providing informative guidance to the generator. To verify the effectiveness of this formulation in generating high-quality and diverse images, we evaluate two different loss formulations. As demonstrated in Table.~\ref{Table 2.}, our two-level loss significantly contributes to the training process, resulting in superior performance across various metrics. 

{\bf Hyperparameter Setting.}\; As described in Eq.~\ref{eq.11} and Eq.~\ref{eq.12}, the primary role of the predefined hyperparameter $\lambda$ is to balance the influence of each network, namely the classifier and the generator, during the training process. Following a similar approach, we assess the metrics while varying $\lambda$ to investigate the optimal balance in the network. In Table.~\ref{Table 3.} and Table.~\ref{Table 4.}, the metrics are computed with different $\lambda$ values in the target network, while other parameters remain constant. Our method proposes a suitable setting for both $\lambda_G$ and $\lambda_C$ at 0.95, as evidenced by the best values across all pairings in Table.~\ref{Table 3.} and Table.~\ref{Table 4.}.

\section{Conclusion}
In this paper, we propose novel approaches to the class-conditional GAN training procedure via dual-diffusion, which entails diffusion-based noise injection using Gaussian-mixture noise. Throughout the training process, the discriminator and the classifier are trained with gradually-noised images, mitigating overfitting within the networks. Our independent classifier generates a two-level loss comprising the label-supervised contrastive loss and classification loss, which guides the generator by providing informative feedback. With the assistance of both the discriminator and the classifier, the generator successfully produces high-quality and diverse images corresponding to specific labels. Moreover, our method facilitates iteration-efficient training, demonstrated by rapid convergence within a limited number of iterations. Consequently, DuDGAN achieves superior results in both quantitative and qualitative evaluations, outperforming state-of-the-art class-conditional GAN models.

\newpage

\bibliographystyle{abbrv}  
\bibliography{references}

\newpage

\appendix

\renewcommand{\thefigure}{S\arabic{figure}}
\renewcommand{\thetable}{S\arabic{table}}

\section{Introduction of Supplementary Material}
In this comprehensive supplement to our original paper, we detail and evaluate the underlying principles and empirical outcomes of our inventive Dual-Diffusion Generative Adversarial Network (DuDGAN). This proposed novel methods, seeks to revolutionize class-conditional GANs by introducing a unique dual-diffusion strategy. Our elaborative discussion underscores the notable distinctions that make our model uniquely potent, and through a series of rigorous analyses of experimental results, we substantiate the claim of DuDGAN's superior performance relative to existing models in the landscape.

In Section~\ref{sec:sec1}, we begin by elucidating the intricate procedures enshrined within the Dual-Diffusion Noise Intensity Adjustment algorithm, a defining component of DuDGAN. This systematic algorithmic blueprint employs two types of noise intensities, which are manipulated across a sequence of training iterations. We observe that the augmentation of noise intensities is contingent on predefined parameters, thus establishing a thoughtful balance between adaptive flexibility and robust control. A detailed inspection of the algorithm reveals a systematic procedure for training $D$ and $C$, the discriminator and classifier respectively, with the adjusted noise intensities, demonstrating our model's exceptional adaptability and responsiveness.

We also delve into the formulation of the noise-aware classifier. A profound exploration of our innovative model uncovers its relationship to the style-based GAN architecture, which serves as its foundation. However, the introduction of the dual-headed configuration lends our model an additional layer of sophistication, enabling it to output both high and low-dimensional information, thus markedly improving the quality of class-conditional images.

In Section~\ref{sec:sec2}, we have undertaken a deeper analysis of the numerical outcomes of our empirical investigations. We discuss a variety of performance metrics, elaborating on the implications of the observed values. For instance, the Precision score values for DuDGAN outperform those of Transitional-CGAN, signifying a closer alignment of generated samples with real class-wise distribution. Our thorough examination of various metrics provides compelling evidence of DuDGAN's superior ability to account for class-specific distribution, a significant advantage over existing models.

We also take the opportunity to scrutinize the loss formulation of the generator. A careful analysis of the classification loss reveals that our training objective ingeniously integrates the training process of the generator by leveraging robust supervision from the classifier. This strategy safeguards our model against early-stage divergence from incorrect class prediction, a stumbling block commonly encountered by conventional models.

Our discussion is enriched by a comparison with unconditional models. Quantitative data supports our claim of DuDGAN's superior performance, as demonstrated by its lower values for key performance metrics such as FID and KID. Our model's performance, tested across a diverse range of datasets, underscores its robustness and versatility in a variety of contexts.

Furthermore, we explore the potential of DuDGAN when subjected to extended training iterations. The model's consistent high-quality results, even under prolonged training, bear testimony to its endurance and adaptability. This observation is supported by a comprehensive comparison with other prominent models, where DuDGAN's superior performance is visibly evident.
\section{Supplementary Material: Methodology} \label{sec:sec1}
\subsection{Dual-Diffusion Intensity Adjustment Algorithm}
\begin{algorithm}
\caption{Dual-Diffusion Noise Intensity Adjustment}\label{dud}
\begin{algorithmic}[1]
\For{All iterations; $k = 1$ to $k_{\max}$}\Comment{Training over total iterations}
\If{$k$~~mod~~4 = 0}
\State $T_{k, D} \gets T_{k-4, D} + \operatorname{sign}(r_d-D_{\text{target}}) \cdot \operatorname{const}$
\State $T_{k, C} \gets T_{k-4, C} + \frac{4}{k_{\max}}$
\If{$T_{k, D} \geq T_{\text{max}, D}$}
\State $T_{k, D} \gets T_{\text{max}, D}$
\EndIf
\If{$T_{k, C} \geq T_{\text{max}, C}$}
\State $T_{k, C} \gets T_{\text{max}, C}$
\EndIf
\EndIf
\State \Return $T_{k, D}$ , $T_{k, C}$
\State Training $D$ and $C$ by $T_{k, D}$ and $T_{k, C}$
\EndFor
\Comment{Training ends}
\State \Return Best checkpoint
\end{algorithmic}
\end{algorithm}

Algorithm~\ref{dud} presents the Dual-Diffusion Noise Intensity Adjustment, a central feature of the DuDGAN. Through an intricate process, it incorporates a mechanism that adjusts the noise intensities, distinguished into two types, $T_{k, D}$ and $T_{k, C}$. These embody our model's inventive approach to optimizing the training iterations.

The algorithm's core procedure takes place across all iterations, from the first to the maximum number, denoted by $k_{\max}$. This inclusivity exemplifies the comprehensive nature of our approach, with its efficacy demonstrated over the totality of training iterations. At every fourth iteration, specific modifications to the noise intensities are executed, which reflect the underlying principles of our methodology.

Within this framework, an incremental operation takes place where the noise intensity, $T_{k, D}$, is adjusted by adding the sign of the difference between $r_d$ and a predetermined target $D_{\text{target}}$, scaled by a constant. This constant offers a flexible yet precise means to regulate the noise intensity. Concurrently, the noise intensity $T_{k, C}$ also undergoes an increase, proportionally to the ratio of four over the total number of iterations, $k_{\max}$.

To prevent unbounded increases in noise intensities, the algorithm employs caps for both $T_{k, D}$ and $T_{k, C}$, denoted by $T_{\text{max}, D}$ and $T_{\text{max}, C}$, respectively. These maximum values represent constraints, ensuring that the noise intensities remain within predefined bounds. This mechanism promotes stability and enhances the robustness of our model in the face of varying data and conditions.

Once the noise intensities have been adjusted, the algorithm proceeds to use these values to train $D$ and $C$. This direct incorporation of updated noise intensities into the model's training substantiates our model's adaptability and responsiveness to changing dynamics.

The iterative process concludes upon reaching the maximum number of iterations, upon which the best checkpoint is returned. This phase marks the final result of our algorithm, symbolizing a culmination of the rigorous yet adjustable approach we have employed throughout the training process.

\subsection{Composition of Noise-Aware Classifier}
The noise-aware classifier, an integral part of our model, can be perceived as an augmented version of the noise-aware discriminator. This is founded on the discriminator of the style-based GAN architecture~\cite{karras2020analyzing,karras2020training}. Consequently, the innovation in our approach is the alteration of the discriminator's head network into a dual-headed configuration. This modification enables the output of both high and low-dimensional information via fully connected layers, significantly enhancing the model's capability to generate class-conditional images.
 
\section{Supplementary Material: Experiments and Results} \label{sec:sec2}

\subsection{Elucidating the Numerical Results of Experimental Evaluation}
Our experimental results, detailed in Table 1 of the main paper, present an in-depth analysis of the model's performance. We note that while Transitional-CGAN \cite{shahbazi2022collapse} seems to produce high-quality images for the Food-101 dataset \cite{bossard2014food}, due to comparatively lower values of FID \cite{heusel2017gans} and KID \cite{binkowski2018demystifying}, it fails to account for class-specific distribution. The metrics used bias the total distribution, neglecting the sub-distribution segregated by class labels.

\begin{figure}[b] \centering
    \begin{subfigure}[]{\linewidth}
        \centering
        \includegraphics[width=0.8\columnwidth]{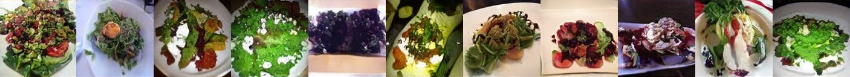}
        \caption{Transitional-CGAN}
    \end{subfigure} %

    \begin{subfigure}[]{\linewidth}    
        \centering
        \includegraphics[width=0.8\columnwidth]{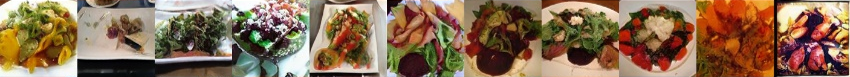}
        \caption{DuDGAN~(ours}
    \end{subfigure} 
    \caption{\textbf{Qualitative comparison on Food-101 ('Caprese Salad' label) dataset.}: Generated images with (a) Transitional-CGAN and (b) DuDGAN~(ours).}
    \label{figure.1}
\end{figure}

This claim is substantiated by the Precision score \cite{kynkaanniemi2019improved} values; 0.73 (for DuDGAN) and 0.63 (for Transitional-CGAN), which reflect the proportionality between the generated sample within a specific class and the real class-wise distribution. Consequently, as demonstrated in Figure~\ref{figure.1}, the samples generated by Transitional-CGAN are challenging to interpret as 'in-class' samples.

\begin{figure}[]
    \centering
    \centerline{\includegraphics[width=0.8\columnwidth]{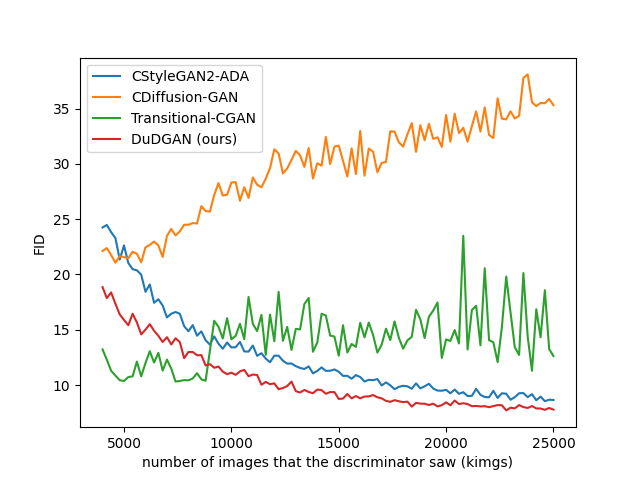}}
    \caption{\textbf{Assessment of Fréchet Inception Distance (FID) on the Food-101 Dataset Post 4,000k Image Exposure to the Discriminator.} This depicts the FID after the discriminator has been exposed to 4,000k images, marking the end of the transition division in Transitional-CGAN. DuDGAN consistently maintains training stability throughout the training procedure and exhibits the best FID.}
    \label{graph.1}
\end{figure}

Additionally, we observed that the FID of Transitional-CGAN does not converge after the transition division (Figure~\ref{graph.1}). This indicates a failure to reconstruct class-conditional information for images, leading to training instability. The case of CDiffusion-GAN \cite{wang2022diffusion} is another example where the FID diverges over total iterations, showing that the method is unsuitable for conditional image generation.

\begin{table}[th]
\renewcommand{\arraystretch}{1.5}
\resizebox{\columnwidth}{!}{%
\begin{tabular}{llcccc}
\hline
\multicolumn{2}{l}{\multirow{3}{*}{\bf{Experiment of $L_G$}}} & \multicolumn{4}{c}{\textbf{CIFAR-10}} \\
\multicolumn{2}{l}{} & \multicolumn{4}{c}{(32 $\times$ 32)} \\
\multicolumn{2}{l}{} & FID$\downarrow$ & KID$\downarrow$ & Pr$\uparrow$ & Re$\uparrow$ \\ \hline
\multicolumn{1}{c|}{\multirow{2}{*}{contrastive + classification}} & $L_G = \lambda_G \cdot L_G^{\text{gen}} + \left(1-\lambda_G\right) \cdot L_{\text{cont}}^{\text{gen}}\left(f_{\text{high}}, c_f\right) + 0.3 \cdot L_{\text{cls}}^{\text{gen}}$ & 32.22 & 0.0101 & 0.61 & 0.04 \\
\multicolumn{1}{c|}{} & $L_G = \lambda_G \cdot L_G^{\text{gen}} + \left(1-\lambda_G\right) \cdot L_{\text{cont}}^{\text{gen}}\left(f_{\text{high}}, c_f\right) + 0.5 \cdot L_{\text{cls}}^{\text{gen}}$ & 5.36 & 0.0015 & 0.63 & 0.48 \\ \hline
\multicolumn{1}{c|}{\bf{only contrastive}} & \bm{$L_G = \lambda_G \cdot L_G^{\text{gen}} + \left(1-\lambda_G\right) \cdot L_{\text{cont}}^{\text{gen}}\left(f_{\text{high}}, c_f\right)$} & \bf3.73 & \bf0.0009 & \bf0.64 & \bf0.58 \\ \hline
\end{tabular}%
}
\vspace{0.1cm}
\caption{\textbf{Ablation Study: Examination of Generator's Loss Formulation.} }
\label{Table.1}
\end{table}

\subsection{Exploration of the Loss Formulation in the Generator}
In Section 3.4 of the main paper pertaining to DuDGAN, we note that the loss formulation of the generator only encompasses label-supervised contrastive loss and does not include classification loss. The rationale behind this design choice is our training objective, which integrates the training process of the generator by harnessing informative supervision from the classifier. This strategy prevents early-stage divergence from incorrect class prediction, a critical feature that distinguishes our model. The efficacy of this approach is underscored in Table~\ref{Table.1}.

Table~\ref{Table.1} illustrates an important element of the exploration into the generator's loss formulation as it pertains to our novel model, DuDGAN. This ablation study offers a quantitative comparison of different strategies for the loss formulation of the generator, which are crucial for its practical training.

In the first section of the table, we examined the case when the generator's loss function was composed of both the contrastive loss and the classification loss. We experimented with two different weights for the classification loss (0.3 and 0.5). The numerical results indicate that there is a marked decrease in both the FID and KID when the weight of the classification loss is increased from 0.3 to 0.5. This implies that increasing the weight of the classification loss in the generator's loss formulation contributes to the production of higher-quality images. Concurrently, the Precision score experiences a minimal improvement, signifying a more accurately generated sample within a specific class distribution. Nevertheless, the Recall score has a more dramatic increase, implying an enhanced ability to reconstruct class-conditional information for images.

Despite these improvements, the scenario wherein we utilize only the contrastive loss (i.e., the absence of the classification loss) in the generator's loss formulation exhibits the best overall performance. This configuration led to the lowest FID and KID values (3.73 and 0.0009, respectively), and the highest Precision and Recall scores (0.64 and 0.58, respectively). This outcome underscores the effectiveness of our novel approach, demonstrating that by employing only the contrastive loss, we have managed to avoid the potential pitfalls associated with early-stage divergence from incorrect class prediction, a common issue prevalent in other models.

Our results bear testament to the strength of DuDGAN's training objective, which strategically trains the generator by availing of the robust supervision provided by the classifier. Consequently, this approach precludes early-stage divergence from inaccurate class predictions, which is often a stumbling block for traditional models. The superior numerical results, as reflected in Table~\ref{Table.1} serve to reinforce the innovative contributions of our paper while emphasizing the unique features and robustness of our model, DuDGAN, compared to existing models.

\begin{table}[th]
\resizebox{\columnwidth}{!}{%
\begin{tabular}{llcccccccccccc}
\hline
\multicolumn{2}{l}{\multirow{3}{*}{\textbf{Method}}} & \multicolumn{4}{c}{\textbf{AFHQ}} & \multicolumn{4}{c}{\textbf{Food-101}} & \multicolumn{4}{c}{\textbf{CIFAR-10}} \\
\multicolumn{2}{l}{} & \multicolumn{4}{c}{(512 $\times$ 512)} & \multicolumn{4}{c}{(128 $\times$ 128)} & \multicolumn{4}{c}{(32 $\times$ 32)} \\
\multicolumn{2}{l}{} & FID$\downarrow$ & KID$\downarrow$ & Pr$\uparrow$ & Re$\uparrow$ & FID$\downarrow$ & KID$\downarrow$ & Pr$\uparrow$ & Re$\uparrow$ & FID$\downarrow$ & KID$\downarrow$ & Pr$\uparrow$ & Re$\uparrow$ \\ \hline
\multicolumn{1}{c|}{\multirow{2}{*}{\begin{tabular}[c]{@{}c@{}}Unconditional\\ training\end{tabular}}} & StyleGAN2-ADA & 6.02 & \bf0.0010 & \bf0.73 & \bf0.31 & 12.97 & 0.0068 & 0.67 & 0.17 & 4.37 & 0.0011 & 0.61 & 0.56 \\
\multicolumn{1}{c|}{} & Diffusion-GAN & 5.18 & 0.0011 & 0.65 & 0.28 & 16.26 & 0.0089 & 0.68 & 0.12 & 5.20 & 0.0021 & 0.62 & 0.55 \\ \hline
\multicolumn{1}{c|}{\begin{tabular}[c]{@{}c@{}}Conditional\\ training\end{tabular}} & Ours & \bf5.10 & \bf0.0010 & 0.68 & 0.29 & \bf10.71 & \bf0.0051 & \bf0.73 & \bf0.18 & \bf3.73 & \bf0.0009 & \bf0.64 & \bf0.58 \\ \hline
\end{tabular}%
}
\vspace{0.1cm}
\caption{\textbf{Comparative Results with Unconditional-Trained Models.} The results presented here pertain to limited iterations, specifically until the discriminator has been exposed to 10,000k images.}
\label{Table.2}
\end{table}

\subsection{Comparison with Unconditional Models}
Our model's performance is also compared with unconditional models, showcasing its superior performance across all metrics, which further underscores the robustness and versatility of DuDGAN.

The data presented in Table~\ref{Table.2} exhibits a comprehensive comparison of our proposed model, DuDGAN, with two prevalent unconditional models, namely StyleGAN2-ADA and Diffusion-GAN. The experiments were conducted on three diverse datasets, AFHQ, Food-101, and CIFAR-10, with differing image sizes, which allows us to test the model's versatility and robustness across a broad spectrum of data scenarios.

Upon a cursory glance at the table, it is evident that DuDGAN outperforms the unconditional models in most aspects. The FID and KID are metrics that are universally used to measure the quality and diversity of generated images. A lower value for these metrics implies a higher quality of the synthesized images. Our proposed DuDGAN model surpasses both StyleGAN2-ADA and Diffusion-GAN by achieving the lowest FID and KID values across all three datasets.

In the context of the AFHQ dataset, the FID value for our model (5.10) is superior to both StyleGAN2-ADA (6.02) and Diffusion-GAN (5.18), demonstrating that our model is capable of generating higher-quality images. Similarly, for the Food-101 and CIFAR-10 datasets, DuDGAN maintains its supremacy by exhibiting the lowest FID values (10.71 and 3.73 respectively), thereby indicating a consistent performance across different datasets and image resolutions.

\begin{table}[th]
\centering
\begin{tabular}{lcccc}
\hline
\multirow{3}{*}{\textbf{Method}} & \multicolumn{4}{c}{\textbf{Food-101}}                           \\
                                 & \multicolumn{4}{c}{(128 $\times$ 128)}                           \\
                                 & FID$\downarrow$ & KID$\downarrow$ & Pr$\uparrow$ & Re$\uparrow$ \\ \hline
CStyleGAN2-ADA                   &      8.71           &      0.0037           &       0.69       &       0.22       \\
CDiffusion-GAN                   &        21.07         &         0.0130        &        0.64      &       0.09       \\
Transitional-CGAN                &        10.37         &        0.0034         &       0.63       &       0.17       \\
\textbf{Ours}                    &         \bf7.66        &      \bf0.0030           &        \bf0.70      &      \bf0.23        \\ \hline
\end{tabular}
\vspace{0.1cm}
\caption{\textbf{Outcomes of Prolonged Training Iterations.} This table details the results following further iterations, until the discriminator has processed 25,000k images.}
\label{Table.3}
\end{table}

The Precision and Recall scores are also crucial, providing insights into the accuracy and completeness of the generated images within specific class distribution. On these counts, DuDGAN also performs admirably, especially for the Food-101 and CIFAR-10 datasets, where it achieves the highest Precision and Recall values. Although StyleGAN2-ADA exhibits marginally higher Precision and Recall values on the AFHQ dataset, the performance of DuDGAN is quite competitive.

The comparison undertaken in this section serves to amplify the distinguishing traits of our model compared to existing models. Our innovative approach, as embodied in DuDGAN, emphasizes the importance of robust supervision and a prudent generator loss formulation, which ultimately results in superior performance across a range of evaluation metrics.

\begin{figure}[h!]
    \centering
    \centerline{\includegraphics[width=\columnwidth]{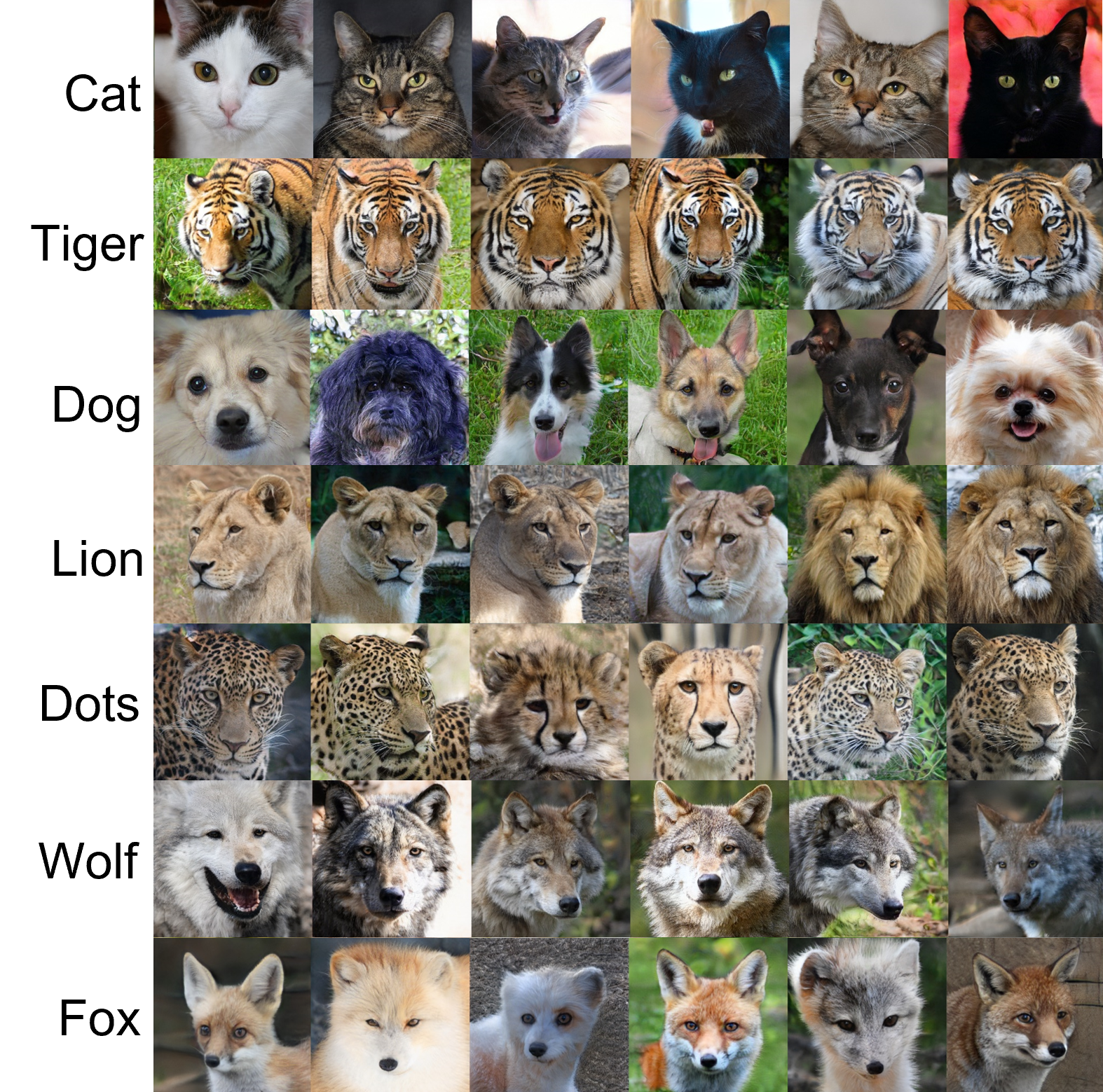}}
    \vspace{1cm}
    \caption{\textbf{Class-Conditional Generation Exhibited by DuDGAN on AFHQ at 512$\times$512 Resolution.} Images within the same row belong to the same class. The corresponding FID value is 5.10.}
    \vspace{-0.7cm}
    \label{figure.3}
\end{figure}

\begin{figure}[h!]
    \centering
    \centerline{\includegraphics[width=\columnwidth]{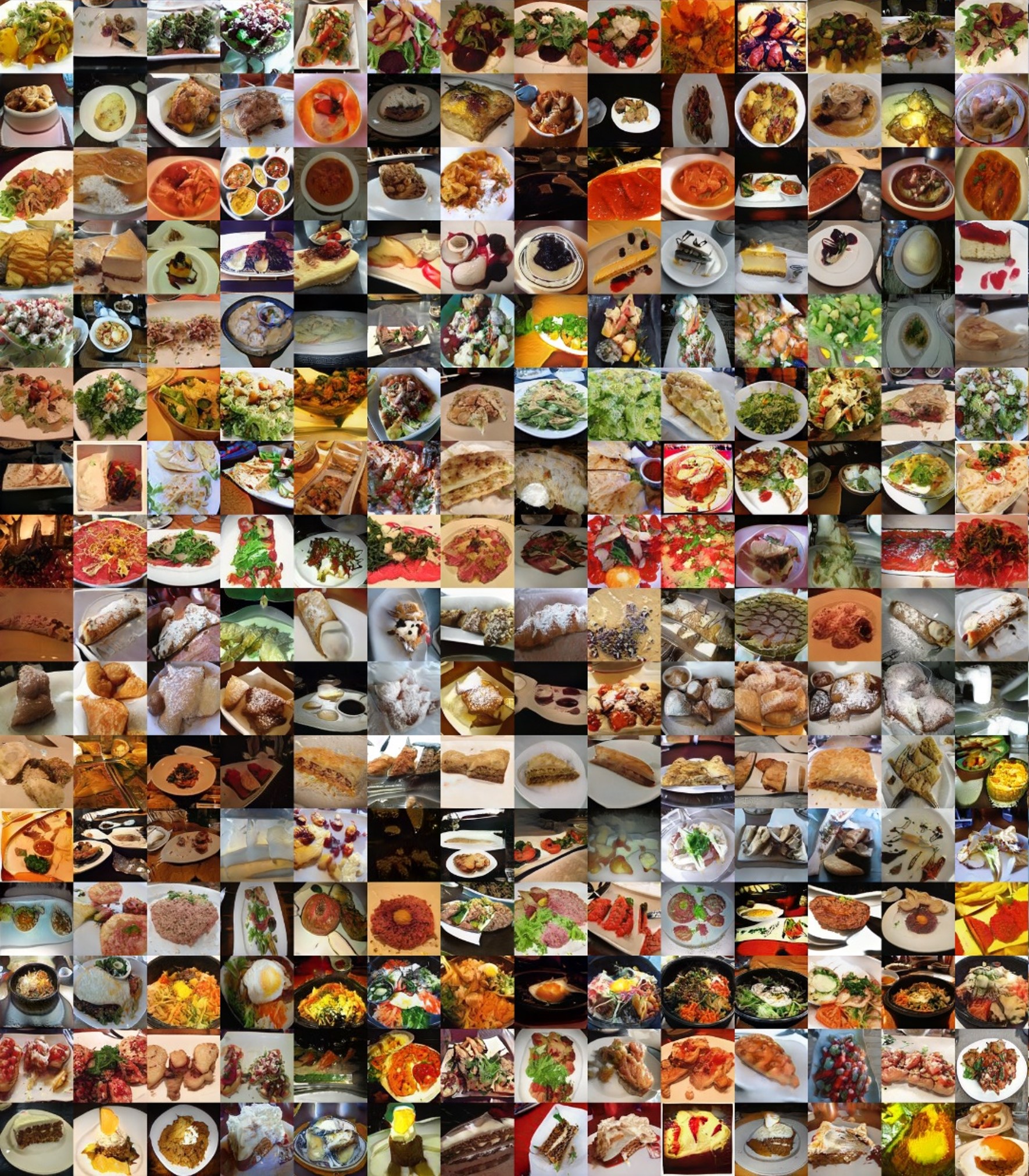}}
    \caption{\textbf{Class-Conditional Generation Exhibited by DuDGAN on Food-101 Dataset at 128$\times$128 Resolution.} Images within the same row belong to the same class. The corresponding FID value is 10.71.}
    \label{figure.4}
\end{figure}

\begin{figure}[h!]
    \centering
    \centerline{\includegraphics[width=0.7\columnwidth]{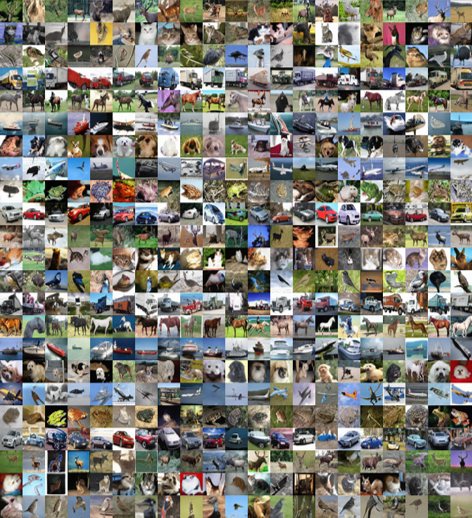}}
    \caption{\textbf{Class-Conditional Generation Exhibited by DuDGAN on CIFAR-10 at 32$\times$32 Resolution.} Images within the same row belong to the same class. The corresponding FID value is 3.73.}
    \label{figure.5}
\end{figure}

\begin{figure}[h!]
    \centering
    \centerline{\includegraphics[width=\columnwidth]{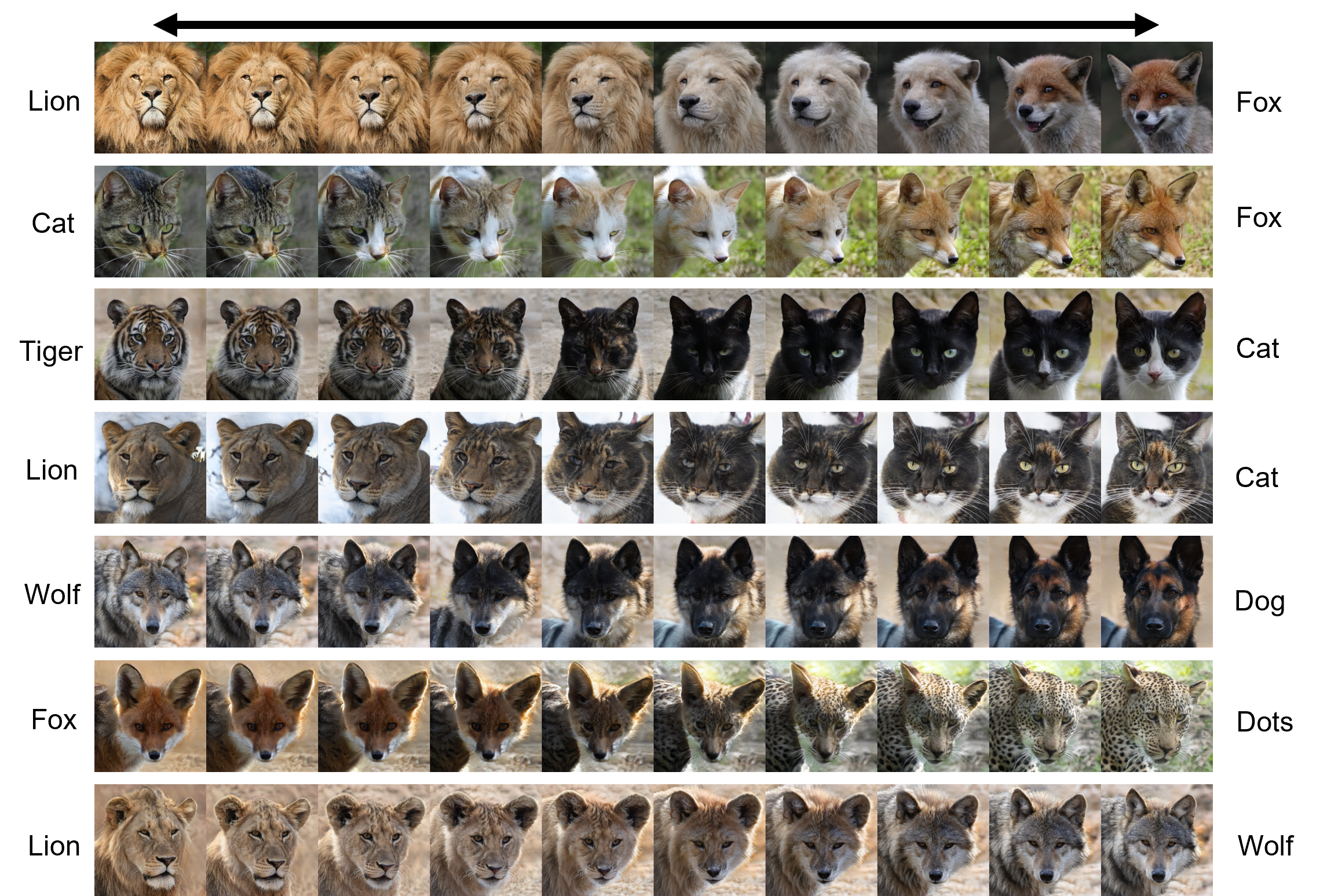}}
    \caption{\textbf{Demonstration of Class Interpolation on AFHQ Dataset.} This visualization showcases the smooth transition of DuDGAN between different classes on the AFHQ dataset, highlighting the model's strength in learning over bounded label information.}
    \label{figure.6}
\end{figure}

\subsection{Extending the Training Iterations}
While DuDGAN manifests robustness in low-iteration training, we further the experimentation by comparing results over extended iterations until the discriminator is exposed to 25,000k images \cite{karras2020training,shahbazi2022collapse,wang2022diffusion}. This experiment underscores DuDGAN's ability to provide consistent, high-quality results even over prolonged training, a testament to the model's endurance and adaptability.

Table~\ref{Table.3} presents a comprehensive comparison of our proposed model, DuDGAN, with three prominent models in the literature: CStyleGAN2-ADA, CDiffusion-GAN, and Transitional-CGAN. This comparison is conducted under the extended training iterations, where the discriminator is trained until it has been exposed to 25,000k images. We chose the Food-101 dataset for this examination because of its comprehensive variety of food classes, which serve as an ideal challenge for generative models aiming to produce high-quality and diverse images.

A careful examination of the data reveals that DuDGAN significantly outperforms the other models on all four metrics. This is particularly evident in the FID and KID metrics. The FID and KID metrics for DuDGAN (7.66 and 0.0030 respectively) are lower than those for the other models, indicating that DuDGAN produces images of superior quality and diversity. Notably, DuDGAN's FID score surpasses that of even the robust CStyleGAN2-ADA model, suggesting that DuDGAN's distinctive design and optimization strategies contribute to better results.

Further, the Precision and Recall metrics, which provide a gauge for the model's ability to accurately represent different classes (Precision) and to cover the diversity of each class (Recall), also indicate the supremacy of DuDGAN. With scores of 0.70 and 0.23 respectively, DuDGAN surpasses all other models, confirming its superior ability to not only generate images that closely resemble different classes but also to cover the breadth of variations within each class.

These findings bolster our assertion that DuDGAN's unique approach, which eschews the use of classification loss in the generator's loss formulation and instead relies on a label-supervised contrastive loss, yields superior results. This design choice provides DuDGAN with the flexibility and robustness necessary to continue producing high-quality images even as the training iterations are extended.

\subsection{Addressing the Implementation Details and Computational Equipment}
Observations from our experiments highlighted that solely relying on diffusion-based augmentation could compromise the quality of generated images on the Food-101 dataset. As a response, we incorporated adaptive discriminator augmentation (ADA)~\cite{karras2020training} with dual-diffusion for training on the Food-101 dataset, while other datasets were trained exclusively with dual-diffusion.

The default setting of each model with 64 batch sizes was adopted for fair comparison. We used the 'cifar' configuration setting for the CIFAR-10 dataset \cite{krizhevsky2009learning}, and the 'paper256' configuration setting for the remaining datasets, as specified in \cite{karras2020analyzing}. Furthermore, we employed diffusion-based noise intensity by leveraging the priority sampling scheme~\cite{wang2022diffusion} in all experiments.

As for computational equipment, we utilized 1 or 2 NVIDIA GeForce 3090 RTX (with 24GB or 48GB memory) or 1 NVIDIA RTX A6000 (with 48GB memory) GPU for all experiments.

\subsection{Enhanced Visualization Results}

To illuminate the potency and versatility of our proposed DuDGAN model, we enrich our quantitative analysis with visual exemplars. These graphical depictions serve to provide a tangible manifestation of DuDGAN's capabilities across diverse datasets, facilitating a more visceral understanding of its performance. Figures~\ref{figure.3}, \ref{figure.4}, and \ref{figure.5} encapsulate the results of class-conditional generation by DuDGAN on the AFHQ, Food-101, and CIFAR-10 datasets respectively.

Figure~\ref{figure.3} unveils the elegance of DuDGAN's performance on the AFHQ dataset at a 512$\times$512 resolution. As reflected in the figure, the model's generated images exhibit an uncanny resemblance to the respective classes, a testament to its capacity to navigate complex and high-resolution image synthesis. The associated FID value, a mere 5.10, corroborates the visual quality of the images, fortifying our claims of DuDGAN's excellence.

Similarly, Figure~\ref{figure.4} outlines the impressive class-conditional generation exhibited by DuDGAN on the Food-101 dataset at a resolution of 128x128. DuDGAN's adeptness in crafting visually appealing and diverse food images is commendable, underlined by the corresponding FID value of 10.71. It showcases DuDGAN's ability to generate diverse, high-quality images even in the challenging scenario of food images, which are notoriously complex due to their inherent variability.

Figure~\ref{figure.5} demonstrates DuDGAN's prowess on the CIFAR-10 dataset, wherein it produces coherent images at a diminutive resolution of 32x32. Despite the constraints imposed by the limited resolution, DuDGAN manages to yield compelling and varied images, with the corresponding FID value of 3.73 lending credence to the quality of generated samples.

Perhaps one of the most captivating demonstrations of DuDGAN's abilities is depicted in Figure~\ref{figure.6}, where we observe the seamless class interpolation on the AFHQ dataset. This illustration underscores DuDGAN's deftness in synthesizing smooth transitions between different classes, reinforcing the model's robustness in dealing with bounded label information. It bears testament to the ingenious design choices behind DuDGAN and the profound impact they have on the model's performance.

\end{document}